\documentclass[conference]{IEEEtran}
\IEEEoverridecommandlockouts

\usepackage{cite}
\usepackage{amsmath,amssymb,amsfonts}
\usepackage{algorithmic}
\usepackage{graphicx}
\usepackage{textcomp}
\usepackage{xcolor}
\usepackage[numbers]{natbib}

\def\BibTeX{{\rm B\kern-.05em{\sc i\kern-.025em b}\kern-.08em
    T\kern-.1667em\lower.7ex\hbox{E}\kern-.125emX}}
\begin{document}
\bstctlcite{IEEEexample:BSTcontrol} 

\title{Random Forest-Supervised Manifold Alignment\\
 \thanks{\textsuperscript{$\dagger$}Both authors contributed equally to this work. This material was supported by a grant from the Simmons Research Endowment.}}

\author{
    \IEEEauthorblockN{Jake S. Rhodes\textsuperscript{$\dagger$}}%
    \IEEEauthorblockA{
        \textit{Department of Statistics} \\
        \textit{Brigham Young University}\\
        Provo, Utah, USA \\
        rhodes@stat.byu.edu
    }
    \and
    \IEEEauthorblockN{Adam G. Rustad\textsuperscript{$\dagger$}}%
    \IEEEauthorblockA{
        \textit{Department of Computer Science} \\
        \textit{Brigham Young University}\\
        Provo, Utah, USA \\
        arusty@byu.edu
    }
}

\maketitle

\begin{abstract}
Manifold alignment is a type of data fusion technique that creates a shared low-dimensional representation of data collected from multiple domains, enabling cross-domain learning and improved performance in downstream tasks. This paper presents an approach to manifold alignment using random forests as a foundation for semi-supervised alignment algorithms, leveraging the model's inherent strengths. We focus on enhancing two recently developed alignment graph-based by integrating class labels through geometry-preserving proximities derived from random forests. These proximities serve as a supervised initialization for constructing cross-domain relationships that maintain local neighborhood structures, thereby facilitating alignment. Our approach addresses a common limitation in manifold alignment, where existing methods often fail to generate embeddings that capture sufficient information for downstream classification. By contrast, we find that alignment models that use random forest proximities or class-label information achieve improved accuracy on downstream classification tasks, outperforming single-domain baselines. Experiments across multiple datasets show that our method typically enhances cross-domain feature integration and predictive performance, suggesting that random forest proximities offer a practical solution for tasks requiring multimodal data alignment. 
\end{abstract}

\begin{IEEEkeywords}
manifold alignment, manifold learning, random forest proximities, multimodal learning, representation learning
\end{IEEEkeywords}

\section{Introduction}

Data collected from an individual observation can come from multiple sources (multi-modal) which may be inherently connected, such as a document translated into multiple languages~\cite{koehn2005europarl}. However, challenges arise when attempting to use multiple data domains in a single model, especially if the data lack clear connections, or when there is no originating source from which both domains are sampled~\cite{wang2011manifold}.

Recent machine learning tools are capable of improving classification and regression problems by incorporating multiple domains in the learning paradigm~\cite{fattahi2023simul-feature-select}. Most recent multimodal models are large and computationally expensive deep learning models~\cite{ramachandram2017deepmultimodalsurvey} which are well formulated for tasks such as generative natural-language processing or image generation~\cite{zhang2020multimodalintelligence}, but are often overkill for smaller, scientific projects. For moderate data sets for scientific tasks, the field of manifold alignment forms a means of generating a shared representation across multiple domains for subsequent tasks.

Manifold alignment aims to find a shared, low-dimensional representation of data sources, enabling the learning of inter-domain relationships while preserving the structure of each domain. Semi-supervised manifold learning involves using known connections, or anchors, to help uncover relationships between domains. In some instances, it is possible to learn richer information by combining insights from multiple sources formed into a single representation, such as a manifold. For example, a classification model can be enhanced by integrating relevant information from multiple data domains. However, we show that many existing manifold alignment methods often yield representations on which predictive models perform poorly compared to single-domain baselines.

In this paper, we overcome this weakness by employing random forests~\cite{breiman2001randomforests} to initialize manifold learning algorithms. We show that using this initialization often improves the accuracy of subsequent supervised models when trained on the resulting joint embedding. We build on recent diffusion- and shortest-path methods~\cite{rhodes2024mashspud}, named, respectively, MASH (Manifold Alignment via Stochastic Hopping) and SPUD (Shortest Path on the Union of Domains). We show that the embedding quality, as measured by post-alignment classification accuracy, is generally improved over existing, semi-supervised alignment methods.

\section{Manifold Alignment Background}

Manifold alignment refers to methodologies aimed at discovering a common, low-dimensional structure, or manifold embedding, that forms a common or shared representation of data from multiple sources. The representation can be subsequently used to analyze relationships across different domains or enhance performance in downstream machine-learning tasks like label prediction. In some cases, alignment is aided by known connections or anchors, represented by data points shared across multiple domains~\cite{wang2011manifold}. In others, alignment relies on discrete, shared data labels~\cite{duque2023mali, tuia2016kema}. These examples belong to a branch known as semi-supervised manifold alignment. Here, we assume that knowledge between points of other modalities or domains can be distilled through known anchor points. At a high level, we assume that point nearness in one modality transfers across domains, extending the idea that ``the neighbor of my neighbors is also my neighbor'' to multiple domains.

This concept was explored in~\cite{rhodes2024mashspud}. In the paper, the authors described two graph-based manifold alignment methodologies that build new graph edges through either diffusion~\cite{coifman2006dm} or shortest paths. They showed that both the diffusion method, Manifold Alignment via Schochastic Hopping (MASH), and the shortest-path method, Shortest Paths on the Union of Domains (SPUD) improved alignment performance compared to state-of-the-art methodologies measured using two independent metrics: The fraction of samples closer than the true match (FOSCTTM)~\cite{liu2019mmd-ma} and cross-embedding classification (CE) or label transfer~\cite{duque2023mali}. We build upon this method by defining weighted graphs via random forest proximities to initialize the graph structure. In doing so, the aligned embedding more frequently improves classification accuracy over that derived from a single domain. In contrast, we show that most manifold alignment methods typically do not enhance the learnable information of classification tasks.

\section{Random Forest Proximities}


A random forest is a supervised learning model designed to predict data labels using an ensemble of randomized decision trees~\cite{breiman2001randomforests}. Each tree is trained on a bootstrap sample, uniformly selected with replacement, with included points referred to as ``in-bag'', and excluded points called ``out-of-bag'' or OOB. 


Similarity measures can be derived from the terminal nodes between the in-bag points and a new observation. Since the similarities are determined by the decision space laid out by the terminal nodes, they can be viewed as a form of supervised similarity. Such similarity measures are often called random forest proximities. Random forest proximities define pairwise relationships between data points, leveraging supervised learning to reveal meaningful patterns guided by the supervision of the random forest model. They have proven to be useful in many applications. One of the original uses of random forest proximities is in data visualization, where they facilitate dimension reduction techniques such as multidimensional scaling (MDS)~\cite{cutler2012randomforests}. Random forest proximities have also been used to improve feature importance measures~\cite{whitmore2018explicating} and data imputation~\cite{Kokla2019metabolomicsimputation}.



The connection between random forest regression models and a nearest-neighbor approach with an adaptive local metric was established in~\cite{lin2006adaptiveNN}, framing the model's predictions in terms of local instance relationships. Building on this concept, Random Forest-Geometry- and Accuracy-Preserving Proximities (RF-GAP)~\cite{rhodes2023rfgap} provided a novel measure that extends this relationship classification contexts and can perfectly replicate random forest out-of-bag predictions. Thus, the RF predictions are described in terms of instance-level attribution; that is, they reflect the influence of adjacent training points. In this way, the model predictions are driven by the importance of local instances in shaping the decision space.

RF-GAP proximities have shown superior performance in proximity-based tasks, including visualization~\cite{rhodes2023supervizedmanifold}, imputation, and outlier detection~\cite{rhodes2023rfgap}. This improvement is likely due to their being intentionally designed to generate random forest OOB predictions, which design also prevents overfitting when used in downstream tasks. By encoding the relationships between data points through random forest proximities, we can establish a meaningful connection between domains through neighboring points.

\subsection{Related Methods}

Manifold alignment methods for semi-supervised learning focus on aligning two modalities or co-domains, $\mathcal{X}$ and $\mathcal{Y}$, where the partial correspondence between points is known. The correspondence is typically defined through points known to belong to both domains (i.e., $x_i$ and $y_j$ form alternative representations of the same point in $\mathcal{X}$ and $\mathcal{Y}$, respectively), although, in some cases, correspondence is inferred via known class labels~\cite{tuia2016kema, duque2023mali}. We briefly describe below existing semi-supervised alignment methods.

Semi-Supervised Manifold Alignment (SSMA)~\cite{ham2005ssma} uses constrained Laplacian Eigenmaps to align known correspondences between domains in low-dimensional embeddings. An alignment method based on Diffusion Maps was derived in \cite{lafon2006datafusion}, performing density normalization across domains and employing an affine transformation to match embeddings. However, limited implementation details preclude us from comparing this approach. Manifold Alignment via Procrustes Analysis (MAPA)~\cite{wang2008ma-procrustes} applies Laplacian Eigenmaps followed by Procrustes alignment on correspondences and generally outperforms SSMA.

Joint Laplacian Manifold Alignment (JLMA) builds on $k$-NN similarities to create a joint Laplacian across domains, embedding known correspondences in a shared coordinate system~\cite{wang2011manifold}. MAGAN~\cite{amodio2018magan}, a Generative Adversarial Network-based approach, aligns domains using reconstruction, discriminator, and custom correspondence losses. Diffusion Transport Alignment (DTA)~\cite{duque2023dta} combines diffusion processes with optimal transport to align domains via cross-domain transition matrices, outperforming over MAGAN, MAPA, and SSMA.

Kernel Manifold Alignment (KEMA)~\cite{tuia2016kema} and Manifold Alignment with Label Information (MALI)~\cite{duque2023mali} are both semi-supervised methods that perform alignment through common data labels rather than known data anchors. KEMA is a kernelized version of SSMA, while MALI uses entropic optimal transport initialized with prior class probabilities, and has been shown to outperform KEMA in label transfer tasks~\cite{duque2023mali}.

\section{Methods}\label{sec:methods}

Both MASH and SPUD~\cite{rhodes2024mashspud} are typically initialized with a kernel matrix for each domain using the $\alpha$-decaying kernel~\cite{moon2019phate}. Instead, we initialize each model using RF-GAP proximities~\cite{rhodes2023rfgap}. The RF-GAP proximities form a row-stochastic diffusion operator which encodes similarities within a given domain. The supervised information, however, is encoded in a way that does not overfit to the training data, which is important for downstream classification tasks. We refer the reader to~\cite{rhodes2023rfgap} for more on this aspect. To learn cross-domain similarity measures, we assign the similarities between anchor points to be maximal. That is, $cross\_similarity(x_i, y_j) = max(p(x_i, x_k))$ for $x_k$ in $\mathcal{X}$, where $p(., .)$ is the normalized proximity, and $y_j$ is the representation of $x_i$ in $\mathcal{Y}$. We form a joint, cross-domain similarity matrix as

\[\mathcal{P} = \left( \begin{array}{cc} P_{\mathcal{X}} & P_{\mathcal{X}\mathcal{Y}} \\ P_{\mathcal{Y}\mathcal{X}} & P_{\mathcal{Y}} \end{array} \right)\]
with $P_{\mathcal{Y}\mathcal{X}}$ = $P_{\mathcal{X}\mathcal{Y}}^T$.

MASH powers the cross-domain similarity, forming new graph connections between the domains. We apply an information distance~\cite{duque2019dig} to the powered diffusion operator, $\mathcal{P}^t$ to perform a potential distance~\cite{moon2019phate} upon which we can apply MDS to form an embedding.

RF-SPUD also starts by forming two RF-GAP similarity graphs, $P_{\mathcal{X}}$, and $P_{\mathcal{Y}}$. Anchor points are then leveraged to form connections between domains. However, instead of following a diffusion process, RF-SPUD learns the shortest paths between the domain graphs by following geodesic curves through anchor points, thereby forming additional cross-domain connections and learning long-range dependencies. A potential distance function~\cite{moon2019phate} is applied to the learned connections and an embedding is formed using MDS. For further details, we refer the reader to~\cite{rhodes2024mashspud}.

\section{Experimental Results}

We investigate the potential for manifold alignment models to produce alignments or embeddings capable of enhancing downstream classification. We compare RF-SPUD and RF-MASH to SSMA, MAPA, JLMA, MAGAN, DTA, KEMA, MALI, and MALI initialized using RF-GAP proximities (RF-MALI). We used 16 classification\footnote{Balance Scale, Breast Cancer, Crx, Diabetes, Ecoli (5 majority classes), Flare1, Glass, Heart Disease, Heart Failure, Hepatitis, Ionosphere, Parkinson's, Seeds, Iris, Tic-Tac-Toe, Cancer Data} files from the UCI Machine Learning Repository~\cite{kelly2024uciml}. We removed observations with missing values and normalized all features from 0 - 1. 


To simulate different modalities from each dataset, we perform feature-level splits (i.e., treat different sets of features as a different domain), and two types of feature distortions. In the \textbf{random} split setup, each domain was assigned a randomly chosen set of features without any overlap. The \textbf{skewed} split assigns domains according to feature relevance as assessed by a random forest model; the most relevant features formed one domain, while the remaining formed the other. For the \textbf{even} split,  features were divided alternating between domains according to importance scores to balance prediction relevance. All remaining less relevant features are assigned randomly.

Two data-distortion domain adaptions were used for comparisons, as presented in~\cite{duque2023dta}. For the first method, \textbf{distort}, Gaussian noise was added to the original features to form a second domain. For the second, random rotations via QR factorization were applied to the original features to produce the \textbf{rotate} domain adaptions.

Our experiments aim to identify alignment methods that produce embeddings with sufficient information to enhance downstream prediction tasks and to determine the contexts in which certain methods perform best. Specifically, we train classification models (random forests and $k$-nearest neighbors) on each domain separately, record validation scores, and then train these models across the aligned, joint embedding space. Ideally, a joint embedding should capture inter-domain relationships, enabling the models to exceed baseline accuracy.

We used a staggered grid search strategy for model parameter selection, testing each parameter set on a fixed 10-nearest neighbor model across each dataset to ensure comparable performance evaluations. We explored combinations of variables with the greatest impact on model performance and reported the best results from the search for each model.

\begin{figure}
    \centering
    \includegraphics[width=\linewidth]{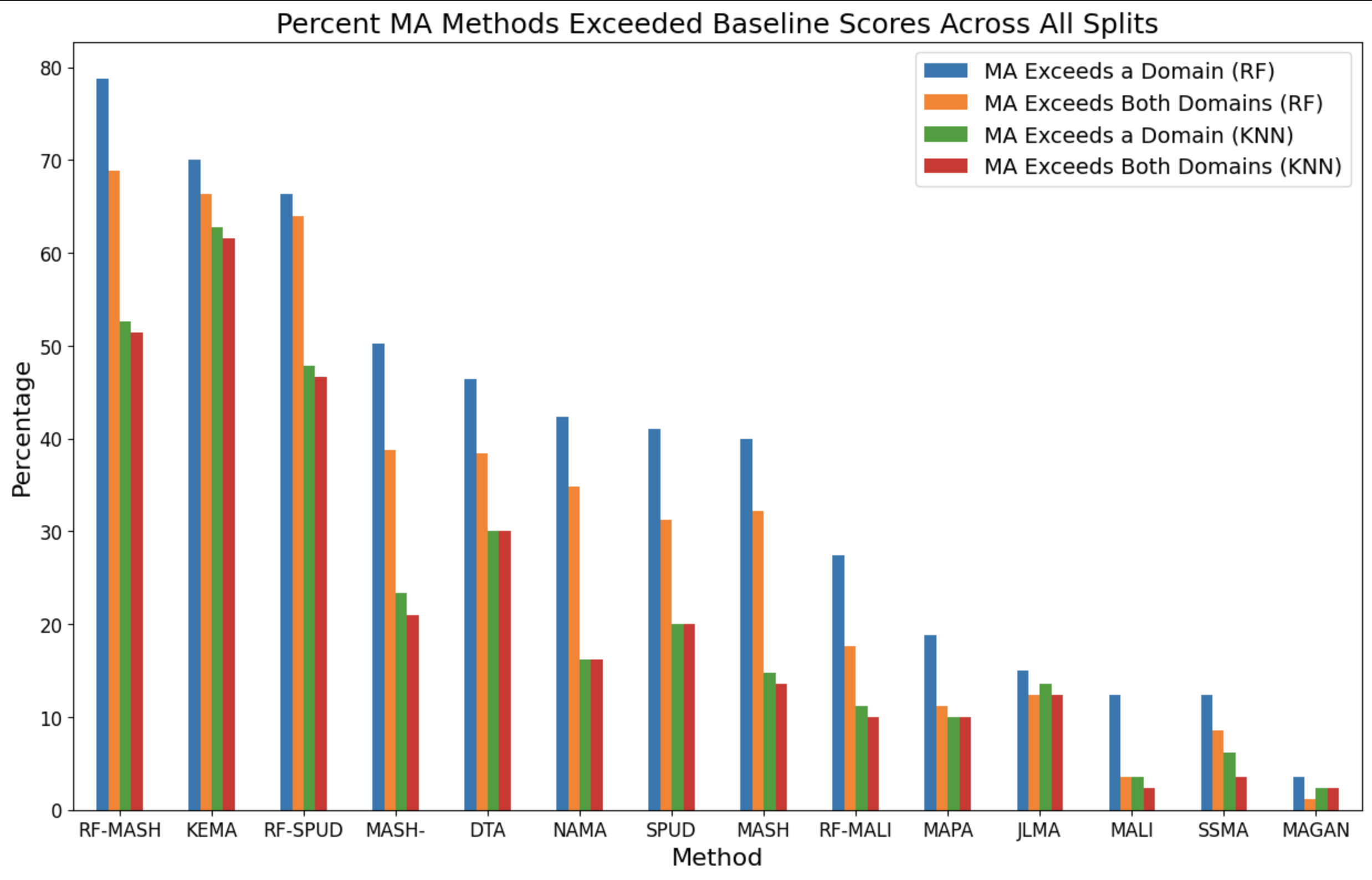}
    \caption{We trained RF and $k$-NN models on each split separately across 5 split types, 16 publicly available datasets, and 3 repetitions each. The models were then trained on the aligned embedding from each method. The proportion of times the classification accuracy exceeded one or both individual models was tracked. Overall, RF-MASH performed best at, exceeding both baselines for random forest models. 30\% of data points served as anchors.}
    \label{fig:better-than-all-splits}
\end{figure}

In Figure~\ref{fig:better-than-all-splits}, we compare baseline classification accuracy with the accuracy of models trained on aligned embeddings. For each dataset and split, we trained a random forest and $k$-NN model, recorded validation accuracy, and then trained each model using manifold embeddings for each alignment method, fitting on one domain and predicting on the other. The proportion of models that outperformed one or both baselines, Overall, RF-MASH, KEMA, and RF-SPUD. Generally, methods initialized with random forests outperformed their unsupervised counterparts.

\begin{figure}
    \centering
    \includegraphics[width=\linewidth]{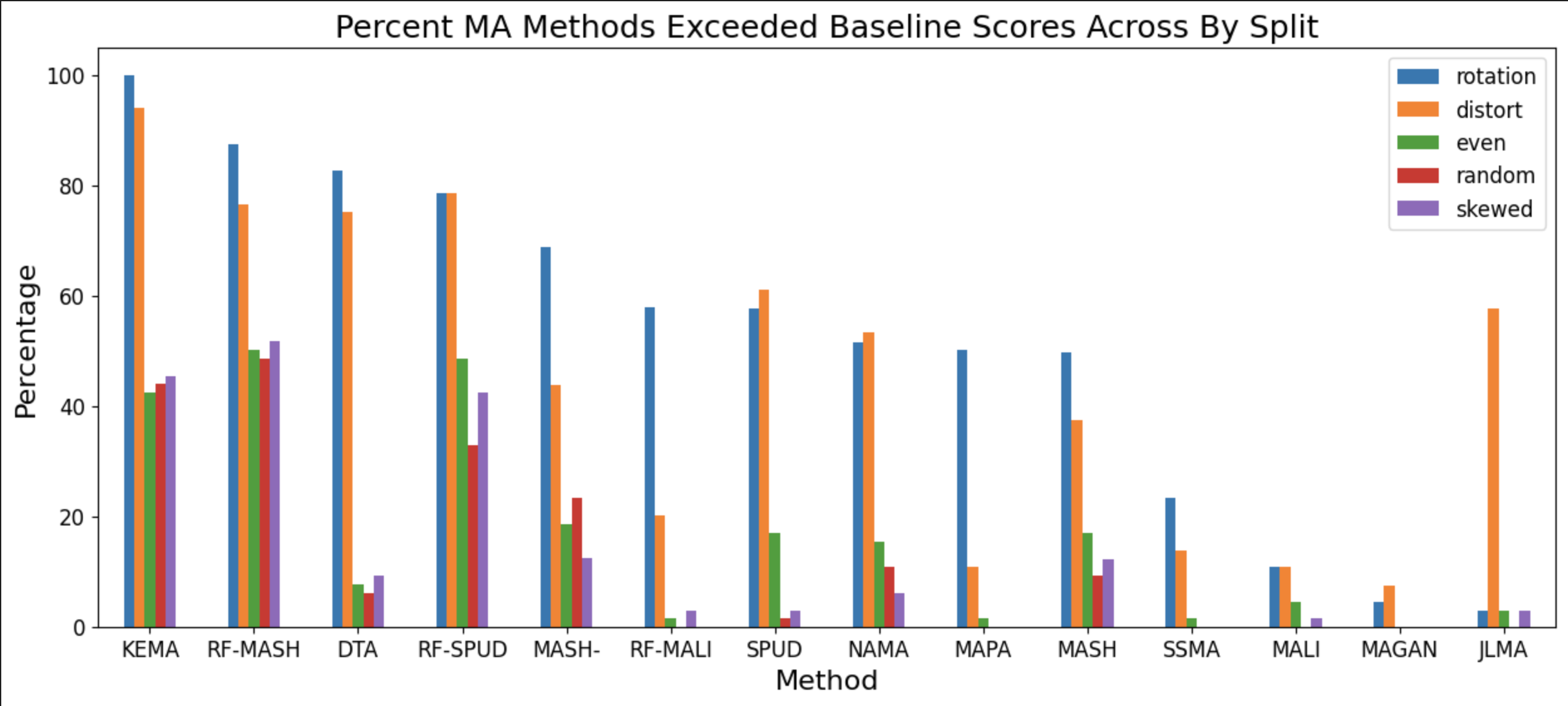}
    \caption{The same datasets and configurations from Figure~\ref{fig:better-than-all-splits} were compared here. In this figure, we distinguish between the different split types used to simulate multiple domains. The RF-initialized methods tend to have more well-rounded results across all split types, though each method fails to exceed 50\% of the baselines for the random and even splits. KEMA performed best at the rotation split but underperformed RF-MASH at all feature-level splits. DTA does well at each distort and rotate, but not well at feature-level splits.}
    \label{fig:better-than-by-split}
\end{figure}

\begin{figure}
    \centering
    \includegraphics[width=\linewidth]{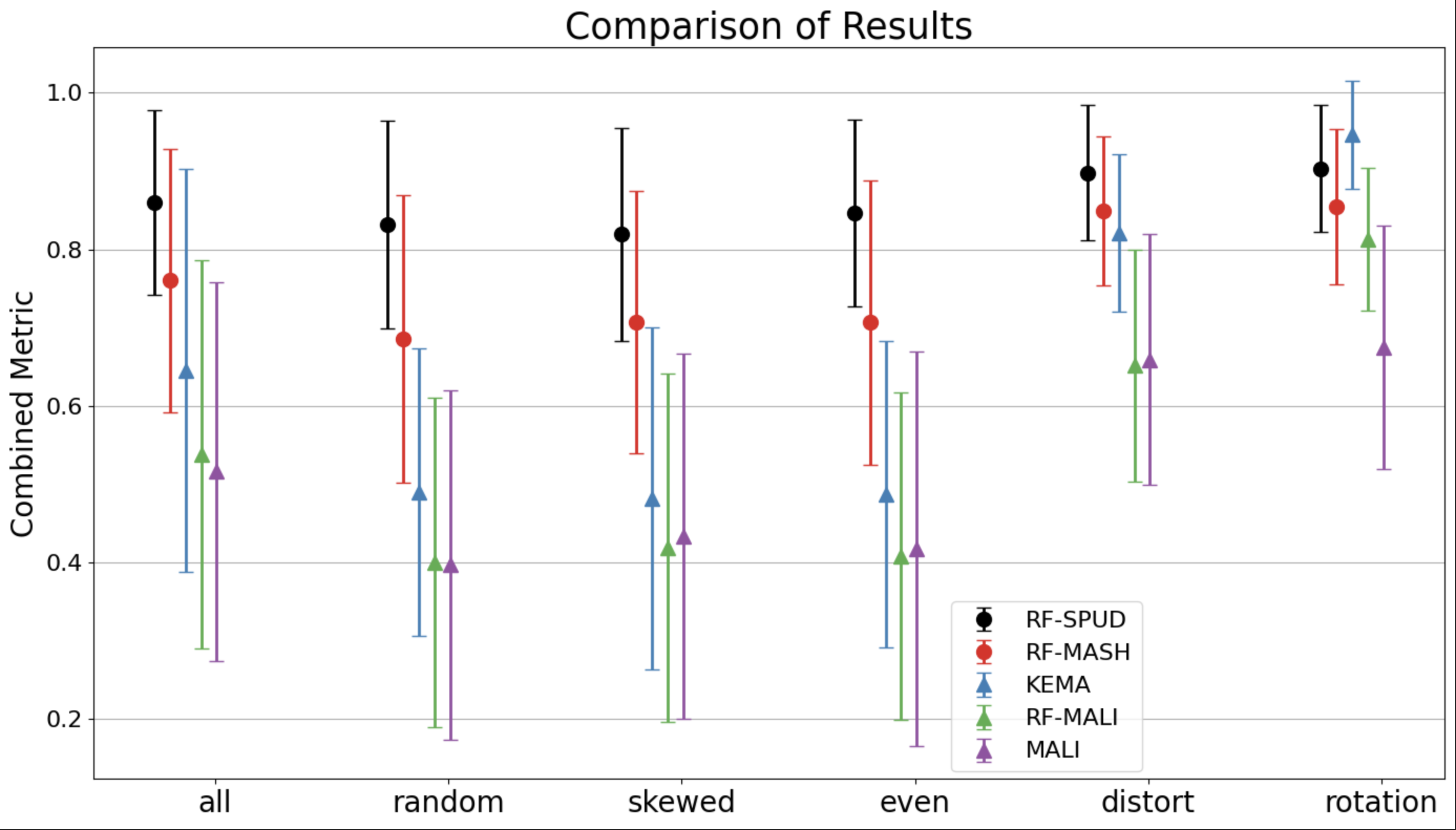}
    \caption{Here we compare all methods that use label information in the alignment process. The combined metric (CE - FOSCTTM)~\cite{rhodes2024mashspud}, is computed and averaged across all datasets. Generally, RF-SPUD and RF-MASH have the best performance with the exception of KEMA on the rotation split.}
    \label{fig:supervised-combined-metric}
\end{figure}

While KEMA generally performs well at improving multi-modal classification accuracy over the baselines, we note that this method underperforms at aligning specific data points to their known correspondences in another domain. For comparison, we include models that use label information during the alignment process using the combined metric (CE - FOSCTTM) as introduced in~\cite{rhodes2024mashspud} in Figure~\ref{fig:supervised-combined-metric}. In this Figure, we see that KEMA is outperformed by RF-SPUD and RF-MASH across all domain adaptations except for random rotations. For all other splits, RF-SPUD and RF-MASH are top performers.

\section{Conclusion}

In this paper, we introduced two semi-supervised alignment methods (RF-SPUD and RF-MASH) initialized using RF-GAP proximities. Although most semi-supervised manifold alignment methods generally do not improve classification models trained on the resulting embeddings, we showed that proximity-initialized models usually enhance aligned manifold embeddings by retaining information sufficient for improving multi-modal classification tasks.




\bibliographystyle{IEEEtran}

{\small  \bibliography{references}}

\end{document}